\pdfoutput=1
\newif\ifpreprintonecolumn
\preprintonecolumntrue

\ifpreprintonecolumn
  \documentclass[journal,onecolumn,11pt]{IEEEtran}
\else
  \documentclass[journal]{IEEEtran}
\fi

\usepackage[numbers,sort]{natbib}
\usepackage{graphicx}
\usepackage{url}
\usepackage{array}
\usepackage{amsmath,amssymb}
\usepackage{paralist}

\ifpreprintonecolumn
  \usepackage[letterpaper,textwidth=5.85in,top=0.75in,bottom=0.80in]{geometry}
\fi

\newcolumntype{L}[1]{>{\raggedright\arraybackslash}p{#1}}

\newenvironment{widefigure}[1][t]{%
  \ifpreprintonecolumn\begin{figure}[#1]\else\begin{figure*}[#1]\fi
}{%
  \ifpreprintonecolumn\end{figure}\else\end{figure*}\fi
}
\newenvironment{widetable}[1][t]{%
  \ifpreprintonecolumn\begin{table}[#1]\else\begin{table*}[#1]\fi
}{%
  \ifpreprintonecolumn\end{table}\else\end{table*}\fi
}

\newlength{\WideFigWidth}
\newlength{\MediumFigWidth}
\AtBeginDocument{%
  \ifpreprintonecolumn
    \setlength{\WideFigWidth}{0.99\columnwidth}%
    \setlength{\MediumFigWidth}{0.86\columnwidth}%
  \else
    \setlength{\WideFigWidth}{0.86\textwidth}%
    \setlength{\MediumFigWidth}{0.76\textwidth}%
  \fi
}
\begin{document}

\title{Aerial Manipulation: Contact, Medium Coupling, and the Geometry of Readiness}

\author{Antonio~Franchi%
\thanks{A. Franchi is with the Robotics and Mechatronics lab, Faculty of Electrical Engineering, Mathematics \& Computer Science, University of Twente, 7522 NB Enschede, The Netherlands, and with the Department of Computer, Control and Management Engineering, Sapienza University of Rome, 00185 Rome, Italy. Email: schol@r-franchi.eu.}
.}

\markboth{Franchi: Aerial Manipulation}{Franchi: Aerial Manipulation}

\maketitle

\begin{abstract}
Aerial robots are increasingly moving from remote observation toward physical interaction with objects, surfaces, structures, loads, and surrounding flows. This review argues that aerial manipulation cannot be understood as classical manipulation simply mounted on a flying base. Because flying agents remain aloft through continuous momentum and energy exchange with the surrounding medium, support, locomotion, stabilization, and task-directed interaction are intrinsically coupled. Building on broad views of manipulation as intentional environmental regulation through physical interaction, we propose a medium-aware interpretation of aerial manipulation in which interaction may be mediated by contact, by the surrounding fluid, or by both. The review organizes biological and robotic examples into a repertoire of interaction modes and a capability ladder, then develops an actuation-geometric viewpoint in which redundancy induces task-equivalent fibers. Internal motion along these fibers can trade energy for active readiness, aerodynamic promptness, and passive medium coupling. This perspective clarifies why aerial manipulation is difficult, why biological flyers remain broader than robotic systems, and how future platforms may command forces while also shaping how the medium acts back on them.
\end{abstract}


\section{INTRODUCTION}

Aerial robots are increasingly leaving the role of remote observers and becoming physical agents: they inspect infrastructure at close range, interact with surfaces, transport objects, deploy tools and sensors, perch on structures, operate through tethers, and cooperate with other robots in carrying or manipulating loads. These developments have produced a rich body of work on aerial manipulation, aerial physical interaction, and aerial robotics for inspection and intervention \cite{olleroAerialManipReview,ruggieroAerialManipReview,meng_he_han,aerialmanipOrsag,ollero-siciliano,Zhong20253528,Zeng20262270}. Yet the conceptual status of ``manipulation'' in the aerial domain remains less settled than the growing number of platforms and applications might suggest. Is aerial manipulation simply classical manipulation performed by a flying base? Or does flight change the nature of manipulation itself?

This review takes the second possibility seriously. It starts from the broad viewpoint that manipulation is not merely the use of a hand or gripper, but a form of intentional environmental control mediated by physical interaction. In this sense, the perspective developed by Mason~\cite{Mason2018TowardManipulation,Mason2001Mechanics}  for robotic manipulation provides an important point of departure: manipulation should be understood broadly, through the coupling between agent, task, body, tools, and environment, rather than only through canonical grasp-and-place tasks. However, flying agents expose a limitation of any view that treats contact as the privileged interaction channel. Aerial systems do not merely move through an environment before interacting with it; they must continuously exchange momentum and energy with the surrounding fluid in order to remain aloft. Their ability to manipulate an object, surface, load, or flow is therefore inseparable from their ability to sustain, stabilize, and reconfigure themselves through the medium.

This observation motivates the central question of this review:
\begin{quote}
\emph{What changes when the manipulator cannot be separated from the medium that makes its action possible?}
\end{quote}
For grounded robots, the support of the agent can often be idealized as given, or at least separated from the manipulation task. For aerial agents, this separation is usually fragile. The same actuation used to generate lift, regulate attitude, reject wind, and maintain flight margins is often also responsible for producing interaction forces, carrying payloads, resisting contact disturbances, or shaping aerodynamic coupling. As a result, locomotion, stabilization, self-support, and task-directed manipulation are not independent modules but different projections of a common physical interaction process.

The goal of this review is therefore not to provide an exhaustive catalog of platforms. Existing surveys and books already describe aerial manipulators, fully actuated multirotors, tethered systems, cooperative load transportation, perching mechanisms, and contact-based inspection platforms \cite{olleroAerialManipReview,ruggieroAerialManipReview,meng_he_han,ref2020-HamUsaSabStaTogFra,ref2017a-TogFra,KumarIJRR11,suarezRALBenchmarking,Dimmig2025153}. Instead, the goal is to propose an organizing language for aerial manipulation: to ask what carries over from classical robotic manipulation, what must be modified when the agent is flying, and what new phenomena become visible only when medium, actuation, and task are considered together.  
Three themes guide the discussion.
\begin{inparaenum}[i)]
    \item First, aerial manipulation requires a broader interaction primitive than selective contact alone: flying agents interact through contact, but also through aerodynamic coupling, wake effects, downwash, inflow, and medium-mediated forces.
    \item Second, aerial actuation has a special structure: part of the actuation budget is continuously spent on self-support, so task authority, disturbance rejection, and energetic cost are coupled.
\item Third, redundancy should be viewed not only as extra actuation, but as an internal geometric resource that can shape readiness, passive response, and interaction capability.  
\end{inparaenum}
The review follows this thread from classical manipulation to aerial definitions, biological and robotic repertoires, actuation geometry, passive interaction, capability taxonomies, and open research directions.  
\section{FROM GROUNDED MANIPULATION TO AERIAL MANIPULATION: WHAT TRANSFERS AND WHAT CHANGES}

A useful starting point for aerial manipulation is to ask which parts of classical robotic manipulation remain valid when the agent is no longer supported by the ground. Mason's broad view of manipulation as environmental control through physical interaction~\cite{Mason2018TowardManipulation,Mason2001Mechanics} remains highly relevant: manipulation is not exhausted by grasping, nor by the motion of a hand-like end-effector, but concerns how an agent uses its body, tools, contacts, and surrounding structures to affect the world.  This viewpoint is particularly valuable for aerial systems because it already weakens the artificial boundary between the robot and its environment. It allows one to regard surfaces, tools, fixtures, contact points, and even parts of the surrounding world as functional elements of the manipulation process.

Several of Mason's central conclusions transfer almost unchanged to flying agents.
\begin{inparaenum}[i)]
\item First, manipulation remains a problem of agent--environment coupling. An aerial robot that pushes on a wall, transports a load, inspects a surface with a contact sensor, perches on a structure, or deploys a tool is regulating a physical interaction, not just executing a trajectory. 
\item Second, the contrast between biological breadth and robotic specialization remains visible in the aerial domain. Flying animals show broad adaptive repertoires, whereas robotic systems often achieve excellent performance in narrower regimes through specialized morphology, sensing, control, and task design~\cite{Alexander2002AnimalLocomotion,DickinsonFarleyFullKoehlKramLehman2000AnimalsMove,DickinsonLehmannSane1999InsectFlight,Sane2003InsectFlightAerodynamics,Dial2003WAIR,KleinHeerenbrink2022AvianPerching,olleroAeroarms,kamel2018voliro,ref2015-RylBueRob,ref2018e-TogFra,Hameed2024,RafeeNekoo20252305,Hameed20258642,Peng2025}.
\item Third, the diagnosis that manipulation is hard remains valid. In aerial systems, perception, planning, contact mechanics, uncertainty, and embodiment are still central difficulties; they are simply coupled to flight dynamics, actuation limits, and energy constraints.
\end{inparaenum}

The locomotion--manipulation duality also transfers, but in a strengthened form. In grounded systems, locomotion and manipulation can sometimes be separated by modeling convention: the base moves the robot, the arm manipulates the world. Mason already points out that this distinction is not fundamental, because locomotion relative to a support can also be viewed as manipulation of that support relative to the agent~\cite{Mason2018TowardManipulation,Mason2001Mechanics,Mason1986Pushing,LynchMason1996StablePushing}. For flying systems, this duality becomes even less optional. Flight is produced by continuous momentum exchange with the surrounding fluid. Thus, the flying agent does not first locomote and then manipulate; it continuously interacts with the medium in order to create the conditions under which the present and future manipulation is possible.

This observation marks the first major discontinuity with grounded manipulation. In Mason's formulation, selective contact is the privileged primitive. For many grounded systems this is appropriate: interaction occurs mainly through hands, fingers, tools, fixtures, wheels, legs, or support contacts. For aerial agents, however, essential interaction is often mediated without direct contact. Lift, thrust, attitude control, gust rejection, wake interaction, ground effect, wall effect, downwash, and inflow-dependent forces are all consequences of fluid-dynamic coupling~\cite{jimenez-canoBridge,Zhang20238114,Dong2023,Paolino2025,ref2018d-FraCarBicRyl,Aucone2023}.  These are not merely disturbances superimposed on the real manipulation task. They are part of the physical channel through which the flying agent sustains itself and, in some cases, affects the environment.

A second discontinuity concerns the status of programmed motion. Classical industrial manipulation can often begin from the premise that a mechanism can impose a desired motion on a load attached to a sufficiently stiff base. This premise is much weaker for aerial robots. A flying platform is not a rigidly supported manipulator in free space. Its position, attitude, and interaction forces are mediated by finite thrust, aerodynamic disturbances, actuator dynamics, payload changes, and stability margins. Therefore, the natural primitive for aerial manipulation is not simply programmed motion, but controlled interaction under self-support constraints. The relevant question is not only whether the robot can reach a pose, but whether it can remain a viable flying agent while producing or absorbing the interaction required by the task.

A third discontinuity is that task mechanics cannot be reduced to contact mechanics alone. Contact, friction, impact, compliance, and unilateral constraints remain essential whenever an aerial system touches the world. However, aerial tasks also involve flight mechanics and fluid mechanics at the same conceptual level. The task may depend on the aerodynamic consequences of proximity to a wall, on rotor downwash acting on nearby objects, on wind-driven load motion, on tether dynamics, or on the passive response of the vehicle to air-relative velocity~\cite{jimenez-canoBridge,ref2017a-TogFra,ref2018h-TogGabPalFra,Dong2023,DAntonio20261769,Zhang20238114}. Thus, the mechanics of aerial manipulation is more accurately described as interaction mechanics: the coupled mechanics of contact, flow, actuation, support generation, load transfer, compliance, and morphology.

These differences do not invalidate the classical manipulation viewpoint. Rather, they enlarge it. A flying manipulator cannot always be understood as an agent first supported by the world and then acting on the world through contact. Its support is already an active interaction, the medium is part of the physical substrate, and the actuation used for locomotion is often the same actuation available for manipulation.

\section{WHAT COUNTS AS MANIPULATION FOR A FLYING AGENT?}

The preceding discussion suggests that a definition of aerial manipulation cannot be obtained by simply adding flight to a contact-centered definition of manipulation. Building on Mason's broad view of manipulation as environmental control through physical interaction~\cite{Mason2018TowardManipulation,Mason2001Mechanics}, the aerial case requires one further step: the relevant physical interaction must include contact and the medium-mediated coupling that makes flight viable.  For a flying agent, the surrounding medium is not merely the space in which manipulation occurs. It is also the medium through which support, propulsion, stabilization, disturbance rejection, and sometimes task action are produced. A multirotor generates lift and control moments by accelerating air through its propellers. A bird or flapping robot modifies the surrounding flow by wing motion and body posture. A fixed-wing vehicle exchanges lift, drag, and moments with the air through its speed, attitude, and aerodynamic surfaces. In each case, the agent is continuously coupled to the fluid; without this coupling, it cannot remain an aerial agent. Thus, the medium is not only a disturbance source or modeling residual. It is part of the interaction substrate~\cite{Paolino2025,RafeeNekoo20252305,Hameed2024,ref2020-HamUsaSabStaTogFra}.

All this motivates the following working definition inspired by~\cite{Mason2018TowardManipulation}:
\begin{quote}
\emph{Aerial manipulation is the intentional regulation of objects, environments, or interaction channels by a flying agent through selective physical interaction, where the relevant interaction may be mediated by contact, by the surrounding fluid, or by both.}
\end{quote}
The phrase ``selective physical interaction'' is important. It prevents the definition from becoming so broad that every motion of a flying body is automatically classified as manipulation. A vehicle passively drifting in wind is interacting with the medium, but it is not manipulating in the sense intended here. Conversely, a vehicle that shapes its rotor speeds, wing posture, body attitude, contact state, or tether tension in order to change the state of an object, maintain a desired interaction, regulate a passive response, or prepare for a disturbance is engaging in manipulation in the broader aerial sense. The selectivity lies in how the agent regulates the physical channels available to it.

This definition includes contact while leaving room for other physical channels. Contact-based aerial manipulation remains central: grasping an object, pushing a surface, sliding along a wall, docking, perching, anchoring, cutting, probing, or applying force with a sensor or tool are all manipulation tasks in a familiar sense~\cite{ref2011-MelLinShoKum,Pounds2011,Yadav2026245,Wu2026,Marconi2011,Albers,bodie2020active}. However, aerial agents may also influence the world through medium-mediated interaction. Rotor downwash may displace, stabilize, or disturb objects; aerodynamic suction and entrainment may alter nearby flow or particle motion; proximity to surfaces may change force production through ground, ceiling, or wall effects; and inflow-dependent aerodynamics may change the passive input--output response of the vehicle~\cite{jimenez-canoBridge,Zhang20238114,Dong2023,Paolino2025,Aucone2023,Hu20241136}. These phenomena need not always be useful or deliberately exploited, but when they are regulated for a task, they belong within the manipulation problem.

It is therefore useful to distinguish two coupled layers. The first is \emph{sustentative manipulation}: the regulation of the medium and of the agent's aerodynamic state to generate support, propulsion, stability, and flight capability. The second is \emph{task-directed manipulation}: the regulation of external objects, surfaces, flows, loads, tools, or contact states. The distinction is conceptual, not physical. The same rotor thrust may support the vehicle and determine a feasible contact force; the same body attitude may produce horizontal acceleration and orient an end effector or sensor; the same flow field may enable flight while disturbing, or deliberately shaping, the task.  
This coupling also clarifies the role of passive interaction. In many aerial systems, the response to an environmental perturbation is not determined only by feedback control. It is also shaped by morphology, aerodynamic state, rotor allocation, compliance, tether geometry, or contact configuration. In this sense, aerial manipulation includes not only the active generation of wrenches, but also the shaping of passive input--output maps between environmental perturbations and robot responses. This viewpoint connects aerial manipulation to broader ideas in impedance control and soft robotics, provided that ``softness'' is understood as a property of interaction rather than only of material composition~\cite{ref2012-ForNalMacMar,ref2012-LipRug,ref2014-RugCacSadLip,Wu20244075,Chen202411028,Hu20231441,McArthur2024558}.

The proposed definition is intentionally modest. It does not claim that all aerial robotics is manipulation, nor that contact-based aerial manipulation should be replaced by a fluid-centered view. Rather, it identifies the broader physical structure within which contact-based tasks, aerodynamic effects, self-support, and passive response coexist. Grounded systems are often dominated by contact-mediated interaction; aerial systems typically combine contact and medium-mediated coupling.  
\section{BIOLOGICAL AND ROBOTIC AERIAL MANIPULATION REPERTOIRE}
\label{sec:bio-rob-repertoires}

The breadth of aerial manipulation becomes clearer when the field is organized by interaction modes rather than by platforms. A flying agent may manipulate by carrying an object, perching on a structure, striking or catching prey, shaping a flow, pulling through a tether, maintaining contact with a surface, or coordinating with other agents to move a load. Some of these behaviors resemble familiar manipulation tasks; others are specific to flight because the agent's support, motion, and interaction authority are continuously mediated by the surrounding fluid. This section therefore uses biological and robotic examples not as an exhaustive catalog, but as evidence of the range of physical interaction channels that aerial manipulation must account for.

\begin{widetable}[t]
\caption{Biological and robotic aerial-manipulation repertoire.\\ The table summarizes interaction classes phenomenologically;\\ several tasks combine multiple rows.}
\label{tab:repertoire-comparison}
\begin{center}
\footnotesize
\renewcommand{\arraystretch}{1.2}
\setlength{\tabcolsep}{2pt}
\begin{tabular}{@{}L{0.20\textwidth}L{0.37\textwidth}L{0.35\textwidth}@{}}
\hline
\textbf{Class} & \textbf{Examples / analogues} & \textbf{Main issues} \\
\hline
\emph{Transport and load} &
Bird carrying; suspended or cooperative payloads &
Load--flight coupling; stability; tension \\

\emph{Perching and anchoring} &
Landing, clinging, perching; claws, spines, docking &
Support switching; impact; hybrid constraints \\

\emph{Surface interaction} &
Inspection, probing, cleaning, NDT contact &
Sustained contact; force regulation; localization \\

\emph{Grasping and capture} &
Prey capture; aerial grasping, catching, hooks, nets &
Fast transient; timing; post-contact recovery \\

\emph{Tethered / cable-mediated} &
Nest materials, silk-like filaments; cables, tethers &
Flexible channel; unilateral tension; oscillations \\

\emph{Medium-mediated} &
Wake use, soaring, hovering; downwash, inflow, wall/ground effect &
Flow as disturbance/resource; deliberate regulation \\

\emph{Collective / distributed} &
Social construction; multi-UAV loads, shared frames &
Distributed authority; coordination; internal forces \\
\hline
\end{tabular}
\end{center}
\end{widetable}

\begin{widefigure}[t]
    \centering
    \includegraphics[width=\WideFigWidth]{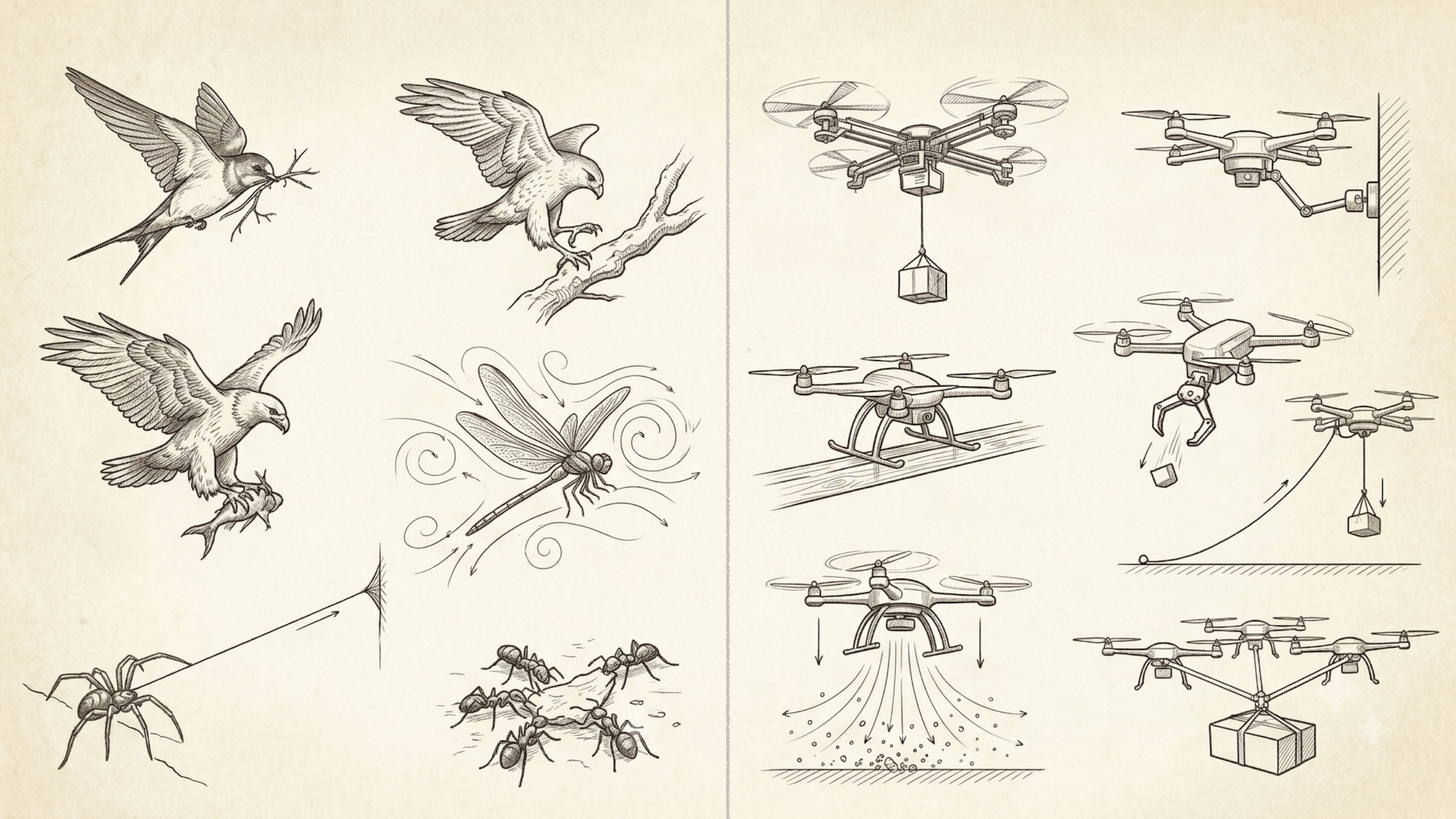}%
    \smallskip
    \caption{Aerial manipulation as a repertoire of interaction modes. 
    Biological flyers and flying or wind-coupled organisms provide examples of flight coupled with carrying, perching, grasping, material interaction, distributed construction, and exploitation of the surrounding medium. 
    Robotic aerial systems reproduce parts of this repertoire through suspended-load transport, contact inspection, perching, aerial grasping, tethered manipulation, flow-mediated interaction, cooperation, and morphology change. 
    The figure is not a taxonomy of species or platforms, but a visual map of interaction channels that motivate the broader definition of aerial manipulation used in this review.
    \emph{(Artistic image by Google Gemini (mid-2026) following the author prompts.)}}
    \label{fig:biological-robotics}
\end{widefigure}

A first class of behaviors is \emph{transport and load interaction}. Bird flight spans a broad repertoire of modes, including take-off, flapping flight, gliding, hovering, soaring, maneuvering, and landing, each imposing different aerodynamic, muscular, and stability requirements~\cite{Videler2005AvianFlight,Tobalske2007BirdFlight,Biewener2022AvianFlight}. Within this repertoire, carrying food, nesting material, or prey is a particularly relevant case for aerial manipulation, because the carried load changes mass distribution, aerodynamic loading, and stability margins during flight. Robotic aerial systems have developed analogous, but typically more specialized, capabilities: single vehicles transporting suspended loads, teams of aerial robots cooperatively carrying objects, and tethered or cable-suspended systems controlling a payload through indirect force transmission~\cite{Michael2011,KumarIJRR11,KumarTRO13,ref2013-SreMicKum,Chai2026116,DeCarli202510546,DAntonio20261769,Oliva-Palomo2024}. In these systems, manipulation becomes the coupled regulation of vehicle dynamics, load dynamics, cable tension, aerodynamic disturbance, and sometimes contact with the environment, rather than only the displacement of a load in space. The object being manipulated is part of the flight dynamics, not an external rigid body moved by a mechanically grounded arm.

A second class is \emph{perching, landing, and anchoring}. Birds, bats, and insects perform rapid transitions between flight and support: they land on branches, cling to surfaces, perch on narrow structures, or hang from ceilings. These transitions are part of a broader repertoire of flight modes and maneuvers, including take-off, flapping, gliding, hovering, soaring, maneuvering, and landing~\cite{Videler2005AvianFlight,Tobalske2007BirdFlight,Biewener2022AvianFlight}. They can also be interpreted as manipulative reconfigurations of the agent--environment coupling: the flying body changes its support condition, exchanges momentum with the environment, and often uses contact to reduce energetic cost or increase sensing and interaction time~\cite{Alexander2002AnimalLocomotion,Dial2003WAIR,BundleDial2003WAIRMechanics,KleinHeerenbrink2022AvianPerching}. Robotic perching systems reproduce parts of this repertoire through claws, spines, adhesives, grippers, impact-absorbing mechanisms, magnetic or suction devices, and morphologies specialized for docking or resting on structures~\cite{shimonomuraIROS2019,Paul2022,Nishio2024660,Kominami202411226,Lee20235027,Scalvini202498,Suarez2023259}. The important structural point is that contact may transform the aerial robot into a hybrid agent. Before contact, the vehicle must generate all support through aerodynamic actuation; after anchoring or perching, the environment may provide reaction forces that change the feasible wrench set, energy demand, and control problem.

A third class is \emph{surface interaction and inspection}. Many aerial robots are designed to approach walls, pipes, tanks, bridges, power lines, or industrial structures while carrying sensors or tools. In contact-based inspection, the platform may press an ultrasonic transducer, camera, cleaning tool, probe, or nondestructive testing sensor against a surface~\cite{ref2016-AleDarBurSie,jimenez-canoBridge,trujilloInspection,bodie2020active,DAngelo2025,Meng20254570,Aucone2025364}. The task is commonly narrow, but technically demanding: it requires stable force regulation, reliable state estimation, contact localization, and robustness to geometry, airflow, and actuator limits. This illustrates a recurring contrast between biology and robotics. Biological flyers often show broad behavioral adaptability, whereas robotic flyers typically achieve high-quality performance by specializing the morphology, sensing, controller, and operational envelope around a restricted class of interactions.

A fourth class is \emph{grasping, capture, and prey-like interaction}. Birds and insects can intercept, strike, grasp, or capture moving targets during fast, visually guided flight, where sensing, maneuvering, and appendage use must be coordinated under severe timing constraints~\cite{DickinsonLehmannSane1999InsectFlight,Sane2003InsectFlightAerodynamics,Combes2012PredatorPreyInteractions,AltshulerSrinivasan2018VisualGuidedFlight,Fabian2018PredatoryFlyInterception,Fowler2009RaptorTalons}. In these behaviors, the body, wings, legs, claws or talons, and sensory systems act together; the manipulation event is inseparable from approach trajectory, aerodynamic maneuvering, target interception, and contact timing.
Robotic analogues include aerial grasping, catching, interception, and object pickup with grippers, nets, hooks, or end-effectors mounted on multirotors or other aerial platforms~\cite{ref2011-MelLinShoKum,Pounds2011,Yadav2026245,Wu2026,Xu20242513,Kumar20243538,Ye20263780,Wang20243656,Indukumar11007808,afifi2026impact}. These systems expose a basic difficulty of aerial manipulation: the interaction often occurs during a fast transient. The robot must arrive with appropriate relative pose and velocity, tolerate impact or contact uncertainty, and maintain enough control authority after the interaction to remain stable. The manipulation event cannot be reduced to a static grasp configuration.

A fifth class is \emph{tethered and cable-mediated manipulation}. Biological systems suggest the broader principle that interaction can be transmitted through appendages, carried materials, and external filaments, rather than only through direct body contact or rigid limbs. Birds routinely manipulate and transport nest materials, and animal builders more generally use external materials---including plant fibers, mud, silk, and other objects---to construct functional structures that extend their interaction with the environment~\cite{Hansell2005AnimalArchitecture,Sheard2023BeakNestMaterialUse}. A particularly clear cable-like biological analogue is provided by spiders, which deploy flexible silk draglines for aerial dispersal; in this case, motion through air is coupled to the mechanics of an external filament interacting with the flow~\cite{Sheldon2017SpiderBallooning}.  In robotics, cables and tethers provide a powerful way to manipulate loads while separating the aerial vehicle from the object or contact site. They also introduce unilateral constraints, tension-only actuation, oscillatory load dynamics, and hybrid contact possibilities~\cite{ref2017a-TogFra,ref2018-Tog,Sanalitro20226726,ref2018h-TogGabPalFra,DAntonio20261769,DeCarli202510546,Das202522337,Sarkisov20235366}. Cable-mediated aerial manipulation is especially important because it expands the manipulator beyond the rigid-arm model. The manipulation channel may be a flexible, underactuated, and dynamically coupled object whose state must be regulated together with the vehicle.

A sixth class is \emph{medium-mediated interaction}. This class has no exact analogue in many grounded manipulation taxonomies. Flying animals exploit the surrounding fluid not only as a support medium, but also as a source of maneuvering authority, stabilization, and energetic opportunity: insects use unsteady mechanisms such as delayed stall, rotational circulation, and wake capture, while birds exploit flight modes involving flapping, gliding, hovering, soaring, take-off, landing, and maneuvering~\cite{DickinsonLehmannSane1999InsectFlight,Sane2003InsectFlightAerodynamics,Videler2005AvianFlight,Tobalske2007BirdFlight}. Similarly, aerial robots may experience or exploit downwash, inflow, ground effect, wall effect, wake interaction, and aerodynamic coupling with nearby objects or surfaces~\cite{Zhang20238114,Dong2023,Paolino2025,Aucone2023,jimenez-canoBridge,Hu20241136}. These effects may be harmful, useful, or both depending on the task: a rotor flow may move lightweight objects, disturb particles, dry or clean a surface, stabilize a flexible element, or alter the local environment. They become manipulation when the agent intentionally regulates them to influence the task or interaction dynamics.

A seventh class is \emph{collective and distributed aerial manipulation}. In biology, the closest analogues are collective construction and material-transport behaviors, especially in social insects, where many individuals coordinate through local interactions, stigmergy, and self-organization to build functional structures~\cite{Hansell2005AnimalArchitecture,PernaTheraulaz2017SocialInsectNests,InvernizziRuxton2019CollectiveBuilding}. Birds provide a complementary example through nest-material manipulation and construction, which couple locomotion, morphology, material handling, and environmental structure~\cite{Hansell2005AnimalArchitecture,Sheard2023BeakNestMaterialUse}. The useful principle is that the effective manipulating agent may be distributed over many bodies and over the material structure being progressively created or modified. Robotic systems similarly use teams of aerial vehicles for cooperative transport, distributed sensing, load stabilization, and multi-point interaction with cables or frames~\cite{KumarIJRR11,Michael2011,KumarTRO13,Chai2026116,Oliva-Palomo2024,DeCarli202510546,Li20235034,Bamert202466,ref2018h-TogGabPalFra,Gabellieri11543151}. These examples show that the boundary of the manipulating agent may itself be distributed. The ``hand'' is not a single end-effector, but a coordinated set of vehicles, cables, flows, and contact points. This reinforces the need for definitions based on interaction structure rather than morphology alone.

Across these classes, the biological--robotic comparison is highly asymmetric. Biological flyers demonstrate breadth, embodiment, and opportunistic adaptation. They combine contact, aerodynamic shaping, compliance, sensing, and fast transitions in ways that remain difficult to reproduce artificially. Robotic aerial systems, in contrast, often achieve their most reliable results by narrowing the task, designing the morphology around a specific interaction, imposing operational structure, or using dedicated sensing and control. This specialization should not be viewed negatively. It is one of engineering's strongest tools. However, it also clarifies the current gap: aerial robots can increasingly perform impressive interaction tasks, but the field still lacks a unified language for comparing contact-mediated, cable-mediated, flow-mediated, and support-mediated manipulation across platforms and species.

The unifying thread across these examples is not a particular morphology or task object. It is the intentional regulation of physical interaction channels---contact, flow, tether, support, morphology, and actuation---by a flying agent.  

\section{INTERNAL ACTUATION GEOMETRY: REDUNDANCY, AERODYNAMIC PROMPTNESS, AND PASSIVE MEDIUM COUPLING}

%\subsection{Redundancy as Task-Equivalent Internal Geometry}

Aerial manipulation is strongly shaped by the internal organization of actuation. In many aerial robots, several actuator configurations can generate the same commanded body wrench, or the same lower-dimensional task variable relevant to flight or interaction. This is usually treated operationally as a control-allocation problem: given a desired force or wrench, one selects actuator commands that satisfy the task while minimizing effort, respecting bounds, or preserving saturation margin~\cite{JohansenFossen2013ControlAllocationSurvey,Durham1992ConstrainedControlAllocation,DurhamBordignonBeck2017AircraftControlAllocation,ref2020-HamUsaSabStaTogFra,ref2018e-TogFra,ref2018-BreDan,kamel2018voliro,Markovic202510232,Li20248770,parkOmniWrench,Bamert202466}. This viewpoint is essential, but incomplete. Redundancy is more than a numerical surplus of actuators over task dimensions: it induces a geometry of task-equivalent internal states. Where the system operates within this geometry affects not only energetic cost, but also how readily the vehicle can respond to future demands.

To make this point precise, consider an internal actuation state \(v\in\mathcal V\) and a task variable \(w\in\mathcal W\). For a multirotor, \(v\) may represent rotor speeds or servo angles, while \(w\) may represent a generated force, moment, or wrench component. More generally, \(v\) may encode propeller speeds, tilt angles, wing variables, morphing coordinates, cable configurations, tensions, or other internal actuation states. The actuation map
\(
    f:\mathcal V \rightarrow \mathcal W
\)
assigns to each internal state the corresponding task output. When the system is redundant, the preimage
\(
    \mathcal F_w = f^{-1}(w)
\)
contains more than one point. This set is the \emph{fiber} of the task value \(w\): it collects all internal states that realize the same task output. The allocation problem is therefore not only ``how to realize \(w\)'', but also ``where on \(\mathcal F_w\) should the system operate?''

\subsection{Aerodynamic Promptness as Active Readiness}

The fiber viewpoint gives a geometric interpretation of control allocation. An energy-minimizing allocator selects a point on \(\mathcal F_w\) according to a cost such as electrical power, aerodynamic power, squared actuator norm, or another effort proxy. Such choices are natural for endurance and thermal management. However, the selected point also determines the local differential map from small actuator variations to small task variations. If actuator rates or accelerations are bounded, two points on the same fiber can have different capacities to generate rapid changes in \(w\). One point may be energetically economical but locally sluggish; another may be more costly but dynamically prepared. This local ability to change the task output rapidly is what is called \emph{aerodynamic promptness}, in analogy with manipulability and dynamic manipulability in robotic mechanisms~\cite{Franchi2026MuscleCoactivation,Franchi2026AeroPromptness,Yoshikawa1985Manipulability,Khatib1987OperationalSpace,Markovic202510232}.

Geometrically, the differential \(df_v\), or a coordinate representation \(J_f(v)\), maps admissible internal rates into feasible task rates. If the internal rates are bounded according to a metric on \(\mathcal V\), this bound is pushed forward through \(df_v\) into a set of task-rate directions. The volume, conditioning, or directional extent of that pushed-forward set measures how readily the system can modify the task output from the current internal state. Promptness is therefore not a new commanded wrench and not an external objective imposed after allocation. It is a local property of the actuation map, the actuator-rate metric, and the selected point on the task fiber. This is particularly relevant in aerial systems, where rotor acceleration limits, motor torque limits, blade inertia, servo limits, tilt dynamics, and aerodynamic response times constrain how quickly a vehicle can modify the forces and moments available for stabilization or interaction~\cite{Franchi2026AeroPromptness,ref2015-RylBueRob,kamel2018voliro,Li20248770,Bamert202466,Hameed20258642,Paul2023460,ref2018-BreDan}. During agile flight, contact, wind rejection, or payload motion, the ability to change force rapidly can matter as much as the ability to generate a steady-state force.

\subsubsection{Fiber Geometry, Co-Contraction, and Energy--Readiness Trade-Offs}

The trade-off between energy and promptness is not purely algorithmic; it is shaped by the geometry and topology of the fiber itself~\cite{Franchi2026MuscleCoactivation}. In a \emph{cooperative regime}, multiple actuators contribute to the task in compatible directions. Increasing the task intensity---for example a force magnitude, wrench magnitude, or transferred power---generally requires increasing the effective contribution of individual actuators in the same direction. The corresponding fiber is then often compact, or at least strongly bounded by the task equation and physical constraints. Energy and promptness may still select different operating points, but the attainable Pareto front remains limited by the bounded internal reallocation.

In an \emph{antagonistic regime}, by contrast, actuators may oppose each other while preserving the same net task output. A low or moderate task intensity can then be realized through high actuator intensities, producing internal aerodynamic loading and increased energetic dissipation without changing the commanded net wrench. The fiber may contain directions along which actuator magnitudes increase together while their net task effect remains fixed. This is the aerial analogue of co-contraction in biological and artificial muscle groups: promptness can be increased by spending additional energy, because the same small actuator-rate variation produces a larger task-rate variation around the internally loaded state~\cite{Franchi2026MuscleCoactivation,FranchiMizzoni2026SignedQuadraticFoliations,ref2018e-TogFra,parkOmniWrench,Markovic202510232,Li20248770,Jia20231930}.

A minimal example is a pair of antagonistic rotors generating a scalar force. If the net force is the difference of two thrust contributions, the same force can be maintained while both rotor speeds increase. The net output remains fixed, but the local sensitivity of force to rotor-speed variations increases. The internal state has moved along a constant-force fiber while increasing both aerodynamic loading and responsiveness. This is structurally analogous to antagonistic muscle coactivation and variable-stiffness actuation, where internal co-contraction can increase readiness or impedance while preserving a task-level equilibrium~\cite{Wolf2016VariableStiffnessReview,BicchiToniettiBavaroPiccigallo2005VSA,Franchi2026MuscleCoactivation,Franchi2026VariableAerodynamicDamping,ref2012-LipRug,ref2020a-NavSabTogPucFra,ref2019e-TogTelGasSabBicMalLanSanRevCorFra,Liu202447134,Bodie20218165}. The analogy is mathematical rather than metaphorical: in both cases, internal motions preserve the task output while changing differential properties that matter for response.

\begin{widefigure}[t]
    \centering
    \includegraphics[width=\WideFigWidth]{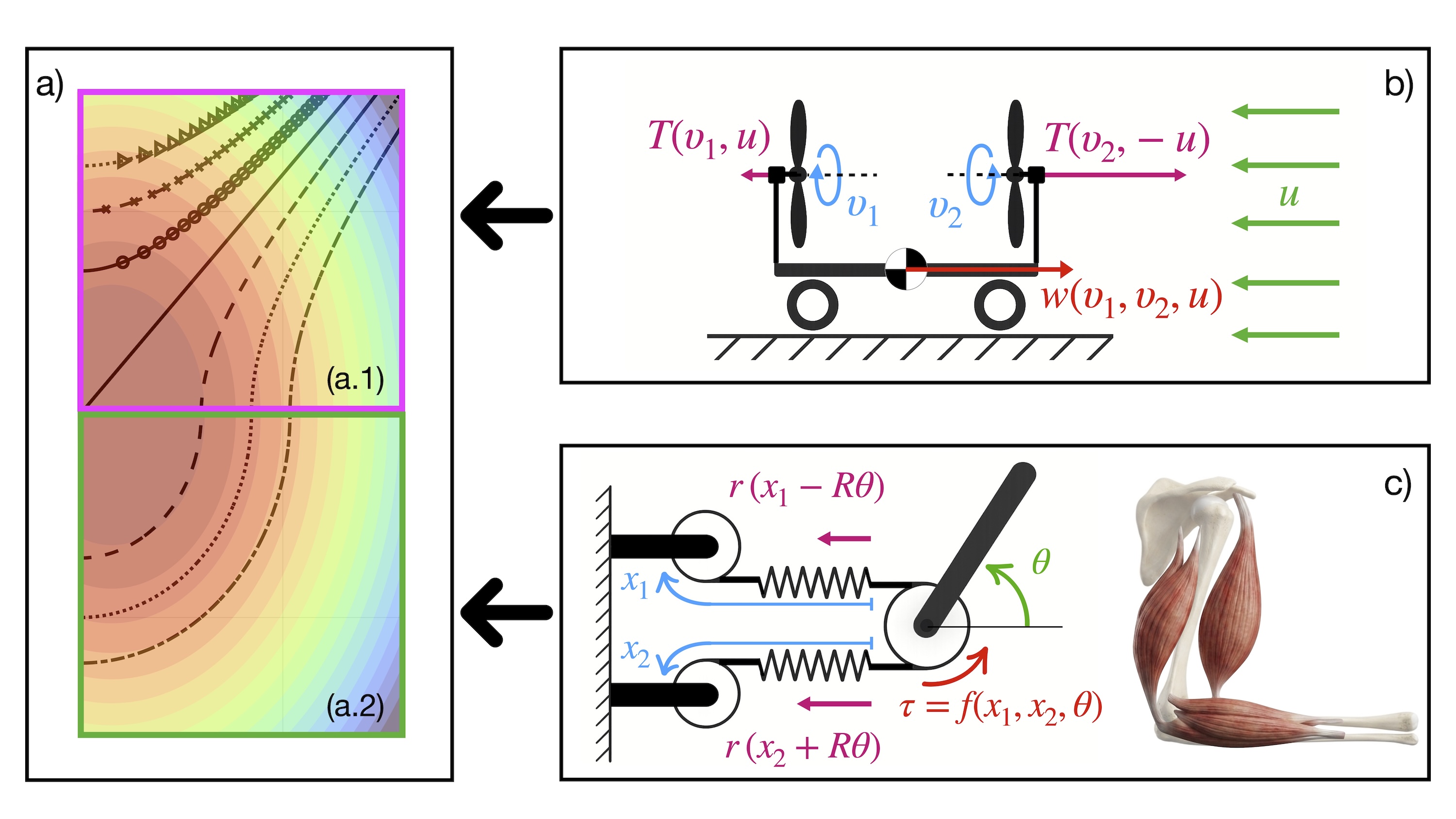}
\caption{Internal actuation geometry, aerodynamic promptness, and the flying-muscle analogy. 
Panel~a) shows task fibers, i.e., level sets of internal states \(v\) realizing the same output \(w=f(v)\), with the promptness cost shown in the background. 
Panels~(a.1) and~(a.2) contrast antagonistic and cooperative regimes: antagonistic fibers allow larger task-equivalent reallocations and broader energy--promptness trade-offs, whereas cooperative fibers are more aligned with cost level curves. 
Panel~b) shows the minimal aerial model: a translating cart with opposed propellers, internal variables \(\nu_1,\nu_2\), axial inflow \(u\), and output \(w(\nu_1,\nu_2,u)\). 
Panel~c) recalls the analogous structure in variable-stiffness actuation and muscle coactivation, with internal variables \(x_1,x_2\) and configuration \(\theta\). 
The common principle is that task-equivalent internal motions can preserve the nominal output while changing local differential properties; in aerial manipulation, this enables redundancy to trade energy for readiness before contact, during load transport, under gusts, or before rapid interaction-wrench changes.}
    \label{fig:aerodynamic-promptness-fibers-vsa}
\end{widefigure}

This interpretation, summarized in
Figure~\ref{fig:aerodynamic-promptness-fibers-vsa}, 
gives redundancy a broader role in aerial manipulation. Beyond avoiding saturation, minimizing energy, tolerating failures, or enlarging the feasible wrench set~\cite{ref2016-AleDarBurSie,jimenez-canoBridge,Dong2023,Paolino2025,Wu20254794,Zhang20238114,Aucone2023}, redundancy can tune readiness. A platform may operate in an energy-efficient allocation during steady flight, then move along a task-equivalent internal direction to increase promptness before contact, during gusty conditions, while carrying a load, or when anticipating a rapid maneuver. Such a change does not alter the nominal task wrench, but it changes the local ability to vary that wrench.

\subsection{Shaping Passive Interaction with the Medium}

The same internal geometry can also affect passive response. The complementary question to promptness is whether the internal state of a flying agent can change the input--output map through which wind, inflow, contact, payload motion, or relative motion affect the vehicle. This question matters because aerial robotics has often treated aerodynamics as something to be modeled, estimated, rejected, or compensated. This view is justified in many settings: wind, propeller interference, ground effect, wall effect, unmodeled drag, and inflow variations can all degrade tracking and interaction performance~\cite{Dong2023,Paolino2025,Zhang20238114,Hu20241136,Wu20254794}. Yet it is not the whole story. The same aerodynamic coupling that appears as a disturbance in one task can become a resource in another. If the robot can modify the physical parameters that determine this coupling, then aerodynamics is not only an external perturbation; it is also an interaction channel that can be shaped.

The antagonistic dual-rotor example also illustrates this passive side. In the static view above, increasing both thrusts while preserving their difference created an internal co-contraction mode that increased active promptness. If thrust also depends on axial inflow, however, the same internal loading changes a passive property of the system. Around a trim velocity, the incremental map from air-relative velocity perturbations to net aerodynamic force perturbations defines an effective aerodynamic damping~\cite{Franchi2026VariableAerodynamicDamping}. Under standard quasi-steady assumptions, this damping increases with the common level of rotor actuation.
Figure~\ref{fig:thickening-air-passive-damping} illustrates the complementary passive interpretation.

\begin{widefigure}[t]
    \centering
    \includegraphics[width=\MediumFigWidth]{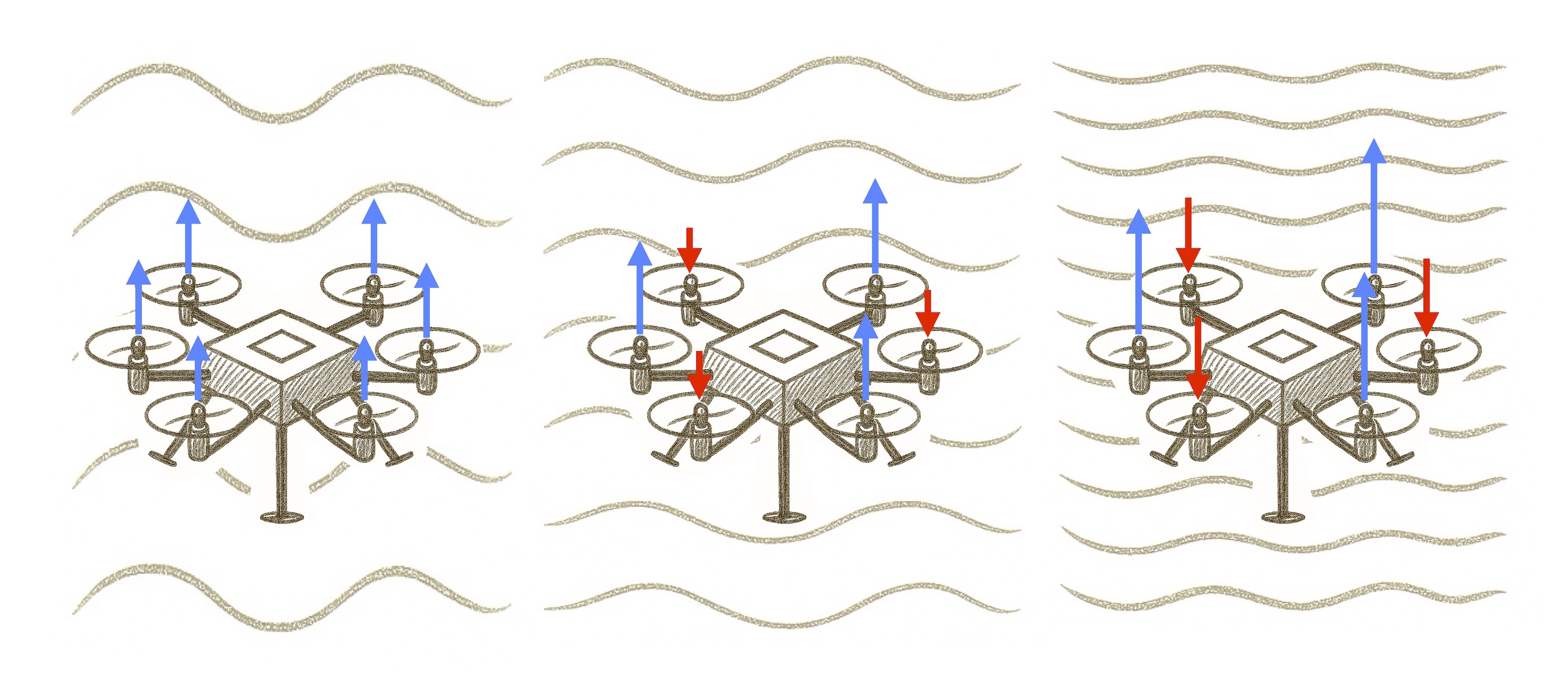}
    \caption{Passive medium shaping through internal actuation. 
The figure illustrates the passive counterpart of Fig.~\ref{fig:aerodynamic-promptness-fibers-vsa}: an antagonistic aerial allocation can preserve the nominal task output while changing the map from air-relative velocity perturbations to aerodynamic force perturbations. 
The depicted ``thickening the air'' is a mechanical metaphor: the vehicle does not change the fluid viscosity, but changes the effective aerodynamic damping or passive impedance around the operating point. 
For aerial manipulation, this suggests that redundancy can be used to select a more favorable passive coupling with the medium before contact, during load transport, near surfaces, or under gusts.}
    \label{fig:thickening-air-passive-damping}
\end{widefigure}

Thus, co-contraction can modulate a passive aerodynamic input--output map while leaving the nominal task output unchanged. The controlled net force may be the same, but the passive response around that operating point is different. This is the aerial counterpart of a familiar principle in variable-stiffness and variable-impedance actuation. In an antagonistic variable-stiffness actuator, internal co-contraction can preserve a nominal joint torque or equilibrium while increasing the passive stiffness seen by external perturbations~\cite{Wolf2016VariableStiffnessReview,BicchiToniettiBavaroPiccigallo2005VSA}. In the aerial case, the corresponding passive quantity need not be stiffness with respect to displacement. Because aerodynamic forces depend naturally on relative flow, the relevant passive map may instead relate velocity perturbations to force variations~\cite{Franchi2026VariableAerodynamicDamping}. The analogy is therefore structural rather than literal: in both cases, an internal actuation variable that is invisible at the task-equilibrium level modifies the local passive interaction experienced by the environment.

For aerial manipulation, this passive viewpoint is not secondary. Contact tasks often fail not only because the desired wrench cannot be produced, but because the robot reacts poorly to small perturbations near contact. Wind gusts, surface proximity, load oscillations, unexpected contacts, and the vehicle's own induced flow can alter interaction forces before feedback control has fully responded. A purely active view asks how the controller should reject these perturbations. A passive-interaction view asks a complementary question: can the operating point, morphology, allocation, or contact configuration be chosen so that the physical robot--environment coupling is already more favorable? Feedback control remains essential, but it acts on top of a passive interaction structure that may itself be designable.

This question is not limited to the dual-rotor example. Similar passive-shaping effects may arise in multirotors with redundant allocation, tilted or variable-pitch propellers, morphing vehicles, winged systems near trim, flapping robots, tethered aerial systems, and platforms that transition between free flight and anchored contact~\cite{Hameed2024,RafeeNekoo20252305,Paolino2025,kamel2018voliro,ref2018e-TogFra,ref2020-HamUsaSabStaTogFra}. In each case, one may ask whether task-equivalent internal states change the passive response to wind, inflow, contact, payload motion, or environmental constraints. The answer will depend on platform physics, operating regime, and the perturbation variables of interest. The important point is not that a universal passive metric already exists, but that the question can now be posed systematically.

\subsection{Active and Passive Axes of Aerial Redundancy}

The concept extends beyond fixed-pitch multirotors, although they provide the clearest initial setting because their actuation maps are nonlinear and their rotor-rate limits have direct physical meaning. Similar questions arise in tilted-rotor systems, variable-pitch platforms, morphing aerial robots, flapping-wing robots, fixed-wing aircraft near trim manifolds, and aerial systems equipped with arms, tethers, or reconfigurable appendages. In each case, one may ask whether the system possesses task-equivalent internal states, how these states are connected, which metric should measure internal rate effort, and how the local task-rate authority or passive response changes along the corresponding fiber. The answers are likely to be platform-dependent: different architectures may require different metrics, different admissible-rate models, and even different notions of promptness or passive impedance. The value of the fiber viewpoint is that it formulates these questions in a common geometric language.

The framework developed in this section gives a precise interpretation of the flying-muscle analogy~\cite{Franchi2026MuscleCoactivation,Franchi2026VariableAerodynamicDamping}. Internal co-contraction can increase active readiness, but it may also change how the vehicle is passively coupled to the medium. Aerial redundancy can therefore be interpreted along two coupled axes. The first is \emph{active promptness}: how rapidly the robot can change its task output under bounded internal rates. The second is \emph{passive impedance}: how environmental perturbations are mapped into robot responses around an operating point. Recognizing both axes enlarges the role of redundancy. Redundancy can satisfy a wrench command, avoid saturation, or minimize power, and it can also shape the physical relation between the flying agent, the surrounding medium, and the task environment.

At present, aerodynamic promptness and passive aerodynamic interaction should be viewed as emerging concepts rather than settled design principles. Their role in aerial manipulation remains to be developed theoretically, tested experimentally, and embedded into allocation, planning, and feedback-control frameworks. Nevertheless, they point to a potentially important shift. Energy, saturation margin, and feasible wrench generation describe what a platform can sustain or produce at a given instant; promptness and passive impedance describe how prepared that platform is to change its interaction with the world, and how it will respond when the world acts back on it. If these trade-offs are better understood, future aerial manipulators may be designed not only to command forces through the medium, but also to shape how the medium acts back on them. Much remains to be done, but this possibility opens a concrete research path for contact, load transport, perching, wind rejection, disturbance recovery, and rapid task transitions.

\section{ELEMENTS OF AERIAL MANIPULATION: A CAPABILITY LADDER}

Aerial manipulation can be organized through interaction capabilities rather than through a single morphology such as a flying arm. In Sec.~\ref{sec:bio-rob-repertoires} we organized the field phenomenologically, by asking which interaction modes appear in biological and robotic systems. The present section reorganizes the same landscape functionally. Its purpose is not to list additional examples, but to identify capability layers that an aerial agent may possess, combine, or transition between. A single task may occupy several layers, and a single platform may be advanced in one layer while remaining simple in another. The ladder is therefore a design and analysis scaffold, not a catalog of behaviors.

\begin{widetable}[t]
\caption{Capability ladder for aerial manipulation.\\ The table abstracts functional layers for analysis and design.}
\label{tab:capability-ladder}
\begin{center}
\footnotesize
\renewcommand{\arraystretch}{1.2}
\setlength{\tabcolsep}{2pt}
\begin{tabular}{@{}L{0.23\textwidth}L{0.34\textwidth}L{0.35\textwidth}@{}}
\hline
\textbf{Layer} & \textbf{Core capability} & \textbf{Relevant metric / design issue} \\
\hline
\emph{Supportive flight} &
Remain viable while acting &
Energy; authority margin \\

\emph{Disturbance-aware flight} &
Approach surfaces, wakes, clutter, payload-induced flow &
Robustness; observability; recovery margin \\

\emph{Contact-capable flight} &
Touch while primarily airborne &
Contact force; compliance; stability \\

\emph{Anchored / perched manipulation} &
Change support condition &
Hybrid stability; reaction forces; energy saving \\

\emph{Grasping and constrained transport} &
Make object, load, or tether part of the dynamics &
Load motion; tension; inertia shift \\

\emph{Aerodynamic manipulation} &
Act deliberately through flow &
Flow authority; disturbance shaping \\

\emph{Cooperative / distributed manipulation} &
Use several agents as one manipulator &
Coordination; internal forces; scalability \\

\emph{Embodied aerial dexterity} &
Change morphology to reshape interaction &
Promptness; passive impedance; energetic cost \\
\hline
\end{tabular}
\end{center}
\end{widetable}

The first layer is \emph{supportive flight}. Every aerial manipulation task presupposes the ability to generate and regulate support through interaction with the medium. For a multirotor, this means producing thrust and moments through propellers; for a fixed-wing vehicle, maintaining lift through forward motion and aerodynamic surfaces; for a flapping system, coordinating wing motion, body dynamics, and unsteady flow~\cite{jimenez-canoBridge,Zhang20238114,Dong2023,ref2016-AleDarBurSie,Paolino2025,Wu20254794}. In aerial manipulation, support is already an active physical interaction, whereas grounded manipulation can often treat support as external to the task. This is the base layer because no subsequent manipulation capability is meaningful unless the agent remains dynamically viable.

The second layer is \emph{disturbance-aware flight near interaction}. Before touching or deliberately affecting the environment, an aerial agent may already be strongly influenced by it. Walls, ceilings, ground proximity, wind gradients, wakes, confined spaces, and payload-induced airflow can alter the effective dynamics of the vehicle~\cite{Albers,Marconi2011,Fumagalli2014,bodie2020active,Meng20254570,Chen20262997,Xu20244419,Aucone2025364,Guo20241576}. At this level, the robot must regulate flight while the environment changes the physical interaction channel, even before a conventional object-manipulation task begins. This layer is important because many failures of aerial manipulation occur before the nominal task begins: the vehicle loses margin, misestimates the flow, saturates actuators, or enters an unfavorable interaction regime.

The third layer is \emph{contact-capable flight}. Here the system intentionally touches the environment while remaining airborne. Examples include pressing a sensor against a wall, sliding along a surface, pushing a button, applying a tool, or maintaining a normal force for inspection~\cite{shimonomuraIROS2019,Paul2022,Nishio2024660,Kominami202411226,Lee20235027,Scalvini202498,Suarez2023259}. The defining feature is the ability to regulate contact while the platform remains primarily supported by aerodynamic actuation, regardless of whether the system carries an arm. This requires force control, compliant behavior, contact-state estimation, and sufficient actuation margin to avoid destabilization. Contact-capable flight is often the first layer that is recognized explicitly as aerial physical interaction, but in the present taxonomy it is only one layer among several.

The fourth layer is \emph{hybrid anchored or perched manipulation}. In this regime, the aerial agent changes its support condition by docking, perching, hooking, bracing, clinging, or otherwise establishing a persistent environmental constraint~\cite{Michael2011,KumarIJRR11,ref2011-MelLinShoKum,Pounds2011,Yadav2026245,Wu2026,DAntonio20261769,Kumar20243538}. This modifies the feasible wrench set and may substantially reduce the energetic cost of maintaining an interaction. It can also improve sensing, allow longer inspection times, and create reaction forces unavailable in free flight. The important conceptual point is that anchoring is not merely stopping flight. It creates a hybrid manipulation mode in which the robot becomes partly aerial and partly environmentally supported. The boundary between locomotion, support, and manipulation is therefore reconfigured.

The fifth layer is \emph{grasping, attachment, and constrained transport}. This includes grasping objects, carrying payloads, attaching to structures, transporting suspended loads, or manipulating through cables and tethers~\cite{Zhang20238114,Aucone2023,Dong2023,Paolino2025,jimenez-canoBridge}. In these tasks, the object or cable is dynamically coupled to the vehicle, and the manipulation problem cannot be separated from flight stability. A suspended load changes the effective dynamics; a tether introduces unilateral tension constraints; a grasped object shifts mass, inertia, and aerodynamic properties. Thus, the manipulated object is not only a task output. It becomes part of the aerial system. This distinguishes aerial transport from many grounded pick-and-place abstractions, where the support of the manipulator is much less directly affected by the carried object.

The sixth layer is \emph{aerodynamic manipulation without direct contact}. Here the vehicle intentionally uses the medium to affect the environment: for example, by using downwash to displace lightweight objects, disturb particles, dry or clean a surface, influence a flexible element, or modify the local flow around another body~\cite{Michael2011,KumarTRO13,KumarIJRR11,ref2018h-TogGabPalFra,Sanalitro20226726,Chai2026116,DeCarli202510546,DAntonio20261769}. This layer must be defined carefully. Not every aerodynamic side effect is manipulation. It becomes manipulation when the flow-mediated effect is deliberately regulated as part of the task. This layer is important because it has no exact counterpart in many classical contact-centered manipulation taxonomies. It makes explicit that aerial agents can act through the medium, not only in the medium.

The seventh layer is \emph{cooperative and distributed aerial manipulation}. Multiple aerial robots can jointly transport loads, regulate cable tensions, stabilize objects, manipulate through shared frames, or distribute sensing and interaction forces over an extended structure~\cite{Jia20231930,Hameed20258642,Scalvini202498,kamel2018voliro,Nishio2024660,Xu20242513,Peng2025,Zhao2025}. In such systems, the manipulating agent is no longer a single body with a single end-effector. It may be a formation, a set of vehicles connected through cables, a distributed actuation system, or a hybrid network of aerial and environmental contacts. This layer highlights that aerial manipulation can be spatially distributed and that the effective ``hand'' of the system may consist of several vehicles, tethers, contact points, and flow fields.

The eighth layer is \emph{morphologically reconfigurable or embodied aerial dexterity}. This includes systems whose body, actuation geometry, wings, propellers, arms, tails, compliance, or appendages change in order to modify the interaction possibilities~\cite{ref2015-RylBueRob,ref2018e-TogFra,ryllInteraction6D,parkOmniWrench,Markovic202510232,Li20248770,Bodie20218165}. Morphology may alter the feasible wrench set, the passive response to perturbations, the ability to perch or grasp, or the energetic cost of interacting. In this layer, manipulation is partly encoded in the body as well as in planned forces or trajectories. This brings aerial manipulation close to embodied design, soft robotics, and biological flight, where morphology and control are deeply intertwined.

This ladder also reveals two axes that cut across all layers: 
The first is \emph{active readiness}: the ability of the agent to rapidly change the task-relevant wrench, motion, or interaction state under actuator limits. This axis is related to promptness and to the geometry of internal redundancy. 

The second is \emph{passive interaction}: the way environmental perturbations are physically mapped into robot responses before, or together with, feedback control. These axes are not additional layers; they are properties that can be present in any layer. A contact-capable inspection robot, a tethered load-transport system, a perched robot, or a morphing aerial vehicle may all be compared not only by what task they perform, but by how much active readiness and passive interaction shaping their architecture permits.

The ladder is intentionally plural. It does not define aerial manipulation by one canonical task, one platform, or one primitive. Instead, it organizes the field around the physical channels through which flying agents regulate their environment: support, disturbance interaction, contact, anchoring, grasping, tethers, aerodynamic action, cooperation, and morphology.

\section{WHY AERIAL MANIPULATION IS HARD}

The broader view developed above also clarifies why aerial manipulation remains difficult. Flying robots are often smaller, lighter, and more disturbance-sensitive than many grounded manipulators, but the deeper issue is that the resources used to remain airborne are often the same resources needed to manipulate. Support, locomotion, stabilization, and task interaction compete for actuation, energy, sensing, and control authority.  

A first source of difficulty is \emph{actuation}. In a grounded manipulator, contact forces are often reacted through a base whose support can be treated as external to the task. In an aerial robot, the actuation used to generate interaction forces is also used to support the vehicle against gravity, reject disturbances, and control attitude. The available wrench for manipulation is therefore a residual quantity: it is what remains after satisfying flight requirements and actuator constraints. This residual character is especially pronounced in underactuated platforms, where translation and attitude are coupled, but it also persists in fully actuated or redundantly actuated platforms because thrust, power, rate limits, and saturation margins remain finite~\cite{jimenez-canoBridge,Zhang20238114,Dong2023,Paolino2025,Hu20241136,Wu20254794}. Thus, a desired interaction wrench must be assessed together with the remaining authority for stabilization, disturbance rejection, and future motion, rather than only as an element of a static feasible set.

A second source of difficulty is the \emph{medium} itself. The air is the support medium, the actuation transmission medium, and a disturbance source at the same time. Wind, turbulence, rotor--rotor interference, downwash, inflow, ground effect, wall effect, and confinement can significantly alter the forces and moments acting on the vehicle~\cite{Lippiello2016,Santamaria-2017,Kim-2016,Guo20241576,Aucone2025364,Chen20262668,Kumar20243538,Guo20241576}. These effects are not necessarily small perturbations around a nominal rigid-body model. Near surfaces, in cluttered spaces, or during contact, the fluid coupling can change precisely when accurate interaction control is most needed. The same physical effect may be harmful, useful, or both: downwash may destabilize an object or be deliberately exploited, and ground effect may disturb height control while modifying the energetic cost of support.

A third difficulty is \emph{perception}. Aerial manipulation often occurs close to surfaces, objects, cables, vegetation, infrastructure, or other robots. These are precisely the situations in which perception becomes harder: cameras may suffer from occlusion, motion blur, poor lighting, or limited field of view; range sensors may be affected by reflective or absorptive surfaces; contact sensors may provide only local information; and aerodynamic effects may not be directly observed~\cite{Marconi2011,pounds2014stability,wopereis2017application,Lee20235027,Nishio2024660,Scalvini202498,Cuniato20226774,Wang20243656}. In addition, the vehicle's own propellers, body motion, and induced flow can disturb the object or environment being sensed. The relevant state is not only the pose of the robot and the object, but also contact state, load motion, tether tension, local flow condition, and sometimes deformation of the environment. This creates a perception problem that is both geometric and physical.

A fourth difficulty is the hybrid nature of \emph{flight--contact transitions}. Many aerial manipulation tasks involve switching between free flight, near-contact flight, impact, sliding, pushing, grasping, anchoring, perching, or tethered interaction. Each transition may change the constraints, the feasible wrench set, the effective inertia, and the stability properties of the system~\cite{afifi2026impact,Indukumar11007808,Jimenez-Cano2013,Markovic202510232,Deng20256661,Meng20254570,Chen20262997,Liang20232774}. These transitions are delicate because they often occur under uncertainty and with limited time for correction. A small error in relative velocity, contact timing, or surface geometry can produce an impulsive load, loss of contact, undesired rebound, actuator saturation, or instability. The challenge is therefore to control each mode and to pass reliably between modes.

A fifth difficulty is \emph{modeling and control}. Aerial manipulation couples rigid-body flight dynamics, actuator dynamics, aerodynamic effects, contact mechanics, load dynamics, and possibly compliance or deformation. Each component may be manageable in isolation, but their interaction creates a system whose relevant model changes with the task. A contact inspection task may require normal-force regulation and surface following; a suspended-load task may require tension constraints and oscillation damping; a perching task may require impact absorption and constraint creation; an aerodynamic manipulation task may require modeling flow-mediated forces~\cite{ref2021-WelPauKum,Deng20256661,Zhang20256027,Li202516405,Housseyn2026119,ref2013-SreMicKum,Li20235034,Dimmig2025153}. Control must therefore be robust to uncertain parameters and changing interaction modes, while remaining fast enough for the time scales of flight.

A sixth difficulty is \emph{planning}. Planning for aerial manipulation is not merely planning a collision-free path to an object. The planner may need to reason about approach direction, aerodynamic disturbance, feasible contact force, wrench margin, payload motion, tether geometry, energy consumption, field-of-view constraints, and safe recovery after interaction~\cite{ref2018h-TogGabPalFra,ref2019b-YueSecBueFra,ref2012-ForNalMacMar,Dong2023,Cuniato20226774,McArthur2024558,Das202522337,Liang20244720}. The resulting problem combines motion planning, task mechanics, medium awareness, and uncertainty. In many cases, the most reliable strategy is not the geometrically shortest or energetically cheapest one, but the one that preserves interaction authority and recovery margins.

A seventh difficulty is \emph{energy}. For aerial agents, energy is not only an optimization criterion; it is a condition for remaining a flying system. Hovering, carrying payloads, rejecting wind, maintaining contact, and increasing promptness all consume energy or reduce endurance~\cite{Cuniato20226774,Liang2026,ref2016-AleDarBurSie,ref2018d-FraCarBicRyl,Deng20256661,ref2020-HamUsaSabStaTogFra,Liang20244720}. This makes aerial manipulation unusually sensitive to the trade-off between efficiency and readiness. An allocation that minimizes energy may leave the vehicle poorly prepared for a disturbance or contact event. Conversely, an internally loaded allocation may increase readiness or passive damping but reduce endurance. The relevant design problem is therefore to decide when energy should be conserved and when it should be spent to buy interaction capability.

Finally, \emph{safety} is structurally central. Aerial robots operate with spinning propellers, stored kinetic energy, limited contact robustness, and potentially fragile surroundings. Interaction failures may damage the robot, the environment, or nearby humans~\cite{SicilianoKhatib2016SpringerHandbook,FloreanoWood2015SmallDrones,RusTolley2015SoftRobots}. Safety is not only a certification layer added after control design; it affects morphology, sensing, planning, actuation limits, interaction strategy, and operational domain.

These difficulties should not be read pessimistically. The same couplings that make the problem hard also reveal structure: actuation redundancy exposes fibers and promptness; medium coupling exposes passive interaction; hybrid support changes feasible manipulation modes; and energy constraints force trade-offs between endurance and readiness.

\section{DIRECTIONS AND OPEN RESEARCH PROGRAM}

The viewpoint developed in this review suggests that more capable platforms and more robust controllers should be accompanied by a structural understanding of how flying agents interact with the medium, the task, and their own internal actuation geometry. The following directions are not a complete agenda, but a set of questions that arise from this coupled view of contact, medium interaction, actuation, and support.
  
A first direction concerns \emph{metrics}. Classical manipulation has benefited from concepts such as manipulability, force closure, task-space inertia, impedance, and grasp quality~\cite{Yoshikawa1985Manipulability,Khatib1987OperationalSpace,Hogan1985ImpedancePartI,BicchiKumar2000GraspingContact,Cutkosky1989GraspChoice,MurrayLiSastry1994Manipulation}. Aerial manipulation requires analogous but medium-aware quantities. For active behavior, one needs metrics that combine feasible wrench magnitude with the ability to change that wrench rapidly under actuator-rate, power, and aerodynamic constraints. This is the role suggested by aerodynamic promptness. For passive behavior, one needs metrics for how perturbations---wind, inflow, contact, payload motion, or surface proximity---are mapped into robot responses around an operating point. A central question is therefore: what metric family can jointly describe active readiness and passive interaction without collapsing platform-specific physics into an overly abstract scalar?

A second direction is the development of canonical models for \emph{medium-coupled task mechanics}. In grounded manipulation, simplified models such as planar pushing, peg-in-hole insertion, rigid grasping, and compliant contact have played an important conceptual role~\cite{Mason2001Mechanics,Mason1986Pushing,LynchMason1996StablePushing,MasonSalisbury1985RobotHands,RaibertCraig1981HybridForcePosition,Hogan1985ImpedancePartI,BicchiKumar2000GraspingContact}. Aerial manipulation would benefit from similarly canonical models: contact near aerodynamic surfaces, rotor-induced flow interacting with objects, load transport under wind, perching transitions, tethered interaction, and trim-dependent passive aerodynamic response. Such models need not capture every detail of fluid mechanics. Their value would be to expose the dominant couplings among actuation, medium, contact, and task, and to clarify which abstractions are safe in which regimes.

A third direction is \emph{morphology--control co-design}. In aerial systems, morphology determines not only reachability or payload capacity, but also the feasible wrench set, the passive response to perturbations, the structure of redundancy, and the energy required for interaction. Rotor placement, tilting mechanisms, variable-pitch propellers, wings, tails, arms, tethers, cages, compliant appendages, and morphing bodies should therefore be analyzed not only as hardware choices, but as design variables that shape the interaction geometry~\cite{RusTolley2015SoftRobots,Wolf2016VariableStiffnessReview,BicchiToniettiBavaroPiccigallo2005VSA,Zeng20262270,Paolino2025,Wu20254794,Wang2026,Li202516405,Ye20263780,Cai2025,Dimmig2025153}. The relevant design question is not simply which platform can generate the largest force, but which architecture provides the right combination of support efficiency, active promptness, passive response, and safe interaction for a class of tasks.

A fourth direction is the integration of \emph{learning with structural insight}. Learning methods are increasingly effective for perception, control adaptation, dynamics modeling, and agile flight~\cite{Marconi2011,Deng20256661,Nishio2024660,Lee2025,Housseyn2026119,ref2021-WelPauKum,Zhang20256027,Li202516405}. However, aerial manipulation is too constrained by safety, data cost, energetic limits, and rare contact events to rely only on unstructured trial-and-error. A promising path is to use learning where models are weak---for example, in flow estimation, contact uncertainty, or residual dynamics---while preserving geometric and physical structure in the representation. The question is not whether learning should replace mechanics, but how learning can be constrained by interaction mechanics, actuation geometry, and invariances that are known before data collection.

A fifth direction is \emph{hybrid planning across flight, contact, and support transitions}. Many aerial manipulation tasks require moving through qualitatively different regimes: free flight, near-surface flight, impact, sustained contact, sliding, grasping, anchoring, perching, tethered motion, and recovery. Planning should therefore reason not only about collision-free paths, but also about when and how the robot changes its support condition, how much wrench margin remains after contact, how energy evolves, and what recovery options exist after failure~\cite{RafeeNekoo20252305,Hameed2024,Ji20258605,Scalvini202498,Aucone2023,Xu20242513,Hameed20258642}. This requires planners that treat contact and medium coupling as task resources, not only as constraints or disturbances.

A sixth direction is renewed comparison between \emph{animals and robots}. Flying animals provide examples of embodied, adaptive, energy-aware interaction with complex environments. Robotic systems provide precise, modular, instrumented platforms through which hypotheses can be tested. The goal should not be imitation for its own sake, but extraction of principles: how support transitions are managed, how morphology stores or dissipates energy, how sensory feedback is distributed, how coactivation trades cost for readiness, and how passive interaction is exploited~\cite{Alexander2002AnimalLocomotion,DickinsonFarleyFullKoehlKramLehman2000AnimalsMove,DickinsonLehmannSane1999InsectFlight,Dial2003WAIR,KleinHeerenbrink2022AvianPerching,RafeeNekoo20252305,Hameed2024,Ji20258605,Scalvini202498,Aucone2023,Xu20242513,Hameed20258642,Peng2025}

Finally, these directions suggest new aerial architectures designed explicitly for \emph{readiness and passive shaping}~\cite{Franchi2026MuscleCoactivation,Franchi2026AeroPromptness,Franchi2026VariableAerodynamicDamping}. Future platforms may not be optimized only for endurance, payload, or hover precision. They may be designed to move along useful internal fibers, to increase promptness when needed, to tune passive aerodynamic damping, to exploit anchoring and perching, or to combine contact and flow-mediated manipulation. The broader research challenge is to understand when such capabilities are worth their energetic and mechanical cost.

The purpose is to make visible structural questions that cut across platforms, rather than to prescribe a single framework. Progress will require theory, design, learning, and experimental systems to develop together.  

\section{CONCLUDING SYNTHESIS}

This review began from a simple question: what changes when the manipulator cannot be separated from the medium that makes its action possible? Classical manipulation theory still applies, but the aerial case exposes a larger interaction structure. A flying agent does not simply move toward a task and then interact with it; it continuously regulates the physical coupling that makes flight possible, and this coupling is often inseparable from the task interaction itself.

Aerial manipulation therefore requires a broadened view of manipulation. Contact remains central, but the surrounding medium also transmits support, disturbances, actuation effects, passive response, and sometimes task-directed action. Actuation likewise generates commanded wrenches while defining internal choices among energy, readiness, saturation margin, failure tolerance, and passive interaction. Redundancy, in this view, is a geometric and physical resource, not merely surplus authority.

The purpose of this viewpoint is to connect existing approaches to aerial manipulation, control allocation, physical interaction, and bio-inspired design through a common language. Concepts such as interaction channels, sustentative and task-directed manipulation, actuation fibers, aerodynamic promptness, co-contraction, and passive medium coupling are useful insofar as they formulate sharper questions and guide better systems. Progress will require precise models, capable mechanisms, robust control, careful experiments, learning where models are incomplete, and continued comparison with biological flight. The challenge is to understand and exploit the aerial nature of the problem.

% Summary Points
\subsection*{Summary Points}
\begin{enumerate}
\item Aerial manipulation involves both contact-mediated and medium-mediated physical interaction: flying agents act on the world while continuously exchanging momentum and energy with the surrounding fluid.
\item Self-support is itself an interaction process, making locomotion, stabilization, and task-directed manipulation less separable in aerial systems than in many grounded systems.
\item Biological flyers display broad, adaptive repertoires of carrying, perching, grasping, material interaction, flow exploitation, and distributed behavior, whereas robotic systems often achieve high performance through task-specific specialization.
\item Actuation redundancy induces task-equivalent internal states that can shape energy use, saturation margin, active readiness, aerodynamic promptness, and interaction capability.
\item Passive aerodynamic interaction can be treated as a design and control object: internal actuation choices may shape how environmental perturbations are mapped into robot responses.
\end{enumerate}

% Future Issues
\subsection*{Future Issues}
\begin{enumerate}
\item A central open problem is to define metrics that jointly quantify feasible wrench generation, active readiness, passive interaction, and energetic cost across different aerial architectures.
\item Canonical models are needed for medium-coupled task mechanics, including contact near aerodynamic surfaces, rotor-induced flow acting on objects, tethered interaction, perching transitions, and trim-dependent passive aerodynamic response.
\item Future aerial platforms should be co-designed across morphology, actuation, and control so that feasible wrench sets, promptness, passive response, and energy consumption are shaped together.
\item Learning methods should be integrated with structural models of actuation, medium coupling, and contact, so that data-driven components improve robustness without discarding known physical invariances and constraints.
\item Further comparison between flying animals and aerial robots may reveal principles for managing support transitions, exploiting passive interaction, trading energy for readiness, and coordinating morphology with control.
\end{enumerate}

%Disclosure
\section*{DISCLOSURE STATEMENT}
The authors are not aware of any affiliations, memberships, funding, or financial holdings that might be perceived as affecting the objectivity of this review. 

% Acknowledgements
\section*{ACKNOWLEDGMENTS}
The author thanks colleagues, mentors, and mentees for inspiring discussions. The author also acknowledges the LLM Gemini 3 (mid 2026) and Copilot (GPT 5.5) for assistance with proofreading, bibliography curation, and image generation. The accuracy of all resulting outputs was carefully and manually verified by the author. The work was partially funded by the European Commission Horizon Europe Framework under project Autoassess (101120732).

\bibliographystyle{IEEEtran}
\bibliography{aerial_manipulation_selected100,aerial_manipulation_older_key60,foundational_musthave}

\end{document}